\documentclass[letterpaper, 10 pt, conference]{ieeeconf}  

\IEEEoverridecommandlockouts                              

\usepackage{graphicx} 
\usepackage{multirow}
\usepackage{float}
\usepackage[utf8]{inputenc}
\usepackage{xcolor}

\title{\LARGE \bf
Tree Species Identification from Bark Images Using Convolutional Neural Networks
}

\author{Mathieu Carpentier, Philippe Giguère and Jonathan Gaudreault}

\begin{document}

\maketitle
\thispagestyle{empty}
\pagestyle{empty}

\begin{abstract}
Tree species identification using bark images is a challenging problem that could prove useful for many forestry related tasks. However, while the recent progress in deep learning showed impressive results on standard vision problems, a lack of datasets prevented its use on tree bark species classification. In this work, we present, and make publicly available, a novel dataset called \emph{BarkNet 1.0} containing more than 23,000 high-resolution bark images from 23 different tree species over a wide range of tree diameters. With it, we demonstrate the feasibility of species recognition through bark images, using deep learning. More specifically, we obtain an accuracy of 93.88\% on single crop, and an accuracy of 97.81\% using a majority voting approach on all of the images of a tree. We also empirically demonstrate that, for a fixed number of images, it is better to maximize the number of tree individuals in the training database, thus directing future data collection efforts.


\end{abstract}

\section{INTRODUCTION}

The ability to automatically and reliably identify tree species from images of bark is an important problem, but has received limited attention in the vision and robotics communities. Early work in mobile robotics has already shown that the ability to recognize trees from non-trees in combined LiDAR+camera sensing can improve localization robustness~\cite{TreeLiDAR2007}. More recent work on data-efficient semantic localization and mapping algorithms~\cite{MatrixPermanent,Toudeshki2018} have demonstrated the value of semantically-meaningful landmarks; In our situation, trees and the knowledge of their species would act as such semantic landmarks. The robotics community is also increasingly interested in flying drones in forests~\cite{SmolyanskiyKSB17}. In terms of forestry applications, one could use this visual species identification to perform autonomous forest inventory. In the context of autonomous tree harvesting operations~\cite{IJFE12426}, the harvester or forwarder would be able to sort timber by species, improving the operator's margins. Similarly, sawmill processes such as debarking could be fine-tuned or optimized based on the species knowledge of the currently processed log. 

For tree species identification, relying on bark has many advantages when compared to other attributes, such as the appearance of its leaves or fruits. First of all, bark is always present despite seasonal changes. It is also present on logs long after the trees have been cut and stored in a lumber yard. In the case of standing tree inventory, bark tends to be visually accessible to most robots, as foliage is not prevalent at the robot's height in forests of commercial value. However, tree species classification using only images of the bark is a challenging task that even trained humans struggle to do, as some species have only very subtle differences in their bark structure. For example, two human experts obtained respectively 56.6\% and 77.8\% classification accuracy on the~\textit{Austrian Federal Forests} (AFF) dataset~\cite{Fiel}.

Recent progress in deep learning have shown that neural networks are able to surpass human performance on many visual recognition tasks~\cite{He2016}. One significant drawback of deep learning approaches is that they generally require very large datasets to obtain satisfactory results. For instance, the ImageNet database contains 14 millions images separated in almost 22,000 synsets.

In the literature, there is no equivalent database for bark recognition, in terms of size or variety. For example, the largest one is the AFF dataset~\cite{Fiel}, with only around 1,200 images covering 11 species. This dataset is also private, making it difficult to use in an open, scientific context. This lack of data might explain why the majority of research on bark recognition has been mostly centered around hand-crafted features such as Gabor filters~\cite{HuangZhi-Kai;Huang2006,1281045}, SIFT~\cite{Fiel} or Local Binary Pattern~\cite{Boudra,SulcSupervisor2014}, as they can be trained using smaller datasets.

To address this issue, we gathered a novel bark dataset specifically designed to train deep neural networks. It contains 23,000 high-resolution images of 23 different tree species found in forests and parks near Quebec City, Canada, from which over 800,000 unique crops of 224x224 pixels can be extracted. The species are typical trees present on the eastern seaboard forests of Canada, most of which have commercial value. In addition to providing the species annotation, we also collected the tree diameter at breast height (DBH), a commonly-used metric in forest inventories. The DBH captures in some sense the age of the tree, thus having the possibility to provide auxiliary information to the network during training. Indeed, bark appearance can change drastically with age, which might help a network optimizer in finding solutions that exhibit better generalization performance. Moreover, having this extra label opens up the possibility to experiment with multi-task learning approaches, for which few datasets exists in the literature~\cite{Zhang2017}.

The contributions presented in this paper are as follow:
\begin{itemize}
\item We collected and curated a novel bark image dataset\footnote{Available at https://github.com/ulaval-damas/tree-bark-classification}, named \emph{BarkNet 1.0}, that is compatible with deep learning research on fine-grained and texture classification problems. This dataset can also be used in the context of multi-task benchmarking.
\item We demonstrated that using this dataset, we can perform visual tree recognition of 20\footnote{For three species, there was an insufficient number of images to perform training and testing.} species, far above any other work. We also quantify the difficulty of differentiating between certain species, via confusion matrices.
\item We performed experiments in order to determine the impact of several key factors on the recognition performance (number of images used during training, use of a voting scheme on classification during testing.)
\end{itemize}

This paper is organized as follows. In Section~\ref{sec:related_work}, we review existing methods and datasets used to accomplish bark image classification. Section~\ref{sec:dataset} introduces our dataset, and details on how it was collected. Section~\ref{sec:experiments} describes the network architecture used to perform classification. Section~\ref{sec:results} presents the results obtained for various test cases. Finally, Section~\ref{sec:conclusion} concludes this paper.

\section{Related work} \label{sec:related_work}

Bark classification has most frequently been formulated as a texture classification problem, for which a number of hand-crafted features have historically been employed. For instance, some works based their approaches on Local Binary Patterns (LBP) \cite{Boudra,SulcSupervisor2014,sulc2013kernel} and others \cite{Fiel} used SIFT descriptors combined with a support vector machine (SVM) to obtain around 70\% accuracy on the AFF dataset. Meanwhile, \cite{Bressane2015} extracted four statistical parameters (uniformity, entropy, asymmetry and smoothness) used in texture classification on trunk images, and employed a decision tree for classification. Furthermore, \cite{Othmani2016} developed a custom segmentation algorithm based on watershed segmentation methods, extracted saliency, roughness, curvature and shape features and fed them to a Random Forest classifier.

Interestingly, some early works used neural networks for bark classification. For instance, \cite{HuangZhi-Kai;Huang2006} extracted texture features based on Gabor wavelet and used a radial basis probabilistic network as the classifier. With their method, they obtained close to 80\% accuracy using a dataset containing around 300 images. This work predates, however, the advent of deep learning approaches, spearheaded by AlexNet \cite{AlexNet}.


With respect to the more general task of tree classification, some did apply deep learning methods. For instance in the \textit{LifeCLEF} competition, which attempts to classify plants using images of different parts such as the leaves, the fruit, or the stem, the best performing methods all employed deep learning~\cite{Champ,Sulc,Sunderhauf,goeau2017plant}. For our purpose however, the number of images with significant bark content in their training database is too small. Less related to the work described herein, work on leaf classification by~\cite{Lee20171} extracted features from deep neural networks, in order to determine what were the most discriminating factors.


Deep learning has also been employed for tree identification from bark information, but using a different type of image. In their work, ~\cite{Mizoguchi} used LiDAR scans instead of RGB images. They used a point cloud with a spatial resolution of 5~$mm$ at a 10~$m$ distance, from which they generated a depth image of size 256x256. For the classification, they fine-tuned a pre-trained AlexNet~\cite{AlexNet} on around 35,000 scans. This allowed them to obtain around 90\% precision on their test set containing 1,536 scans. However, they only used two different species, Japanese Cedar and Japanese Cypress, making the problem significantly less challenging.


Finally, some authors have started exploring deep learning on RGB images of textures. By leveraging extracted features from CNNs pre-trained on ImageNet and different region segmentation algorithms, \cite{Cimpoi_2015_CVPR} used an SVM to classify texture materials, notably on the \emph{Flickr Material Dataset}~\cite{Sharan2009}. They also improved the state-of-the-art by at least 6~\% on all of the datasets on which they tested. Also, \cite{7899932} modified the standard convolutional layer to learn rotation-invariant filters. They did this by grouping filters into groups and by tying the weights of each filter within the same group so that they would all correspond to a rotated version of each other. They tested their layer on the three \emph{Outex}~\cite{mvg:314} texture classification benchmarks and improved the state-of-the-art on one of these benchmarks and obtained similar results on the other two.

\section{Bark dataset (\emph{BarkNet 1.0})} \label{sec:dataset}

\subsection{Existing bark datasets}
One significant hurdle when trying to use deep learning for bark classification is the lack of existing datasets for training purposes. Table~\ref{table:existing_datasets} shows datasets that were used in previous work for the bark classification task. Note that most of these datasets contain only a very small number of images as well as a limited number of classes. Moreover, only one of those datasets is publicly available, hindering the global research effort on this problem.

\begin{table}[t]
\vspace*{3mm}
\renewcommand{\arraystretch}{1.3}
\centering
\begin{tabular}{| c | c | c | c |}
\hline
Reference & Number of classes & Number of images & Public\\
\hline
\cite{1281045} & 8 & 200 & No\\
\hline
\cite{HuangZhi-Kai;Huang2006} & 17 & 300 & No\\
\hline
Trunk12\cite{trunk12} & 12 & 393 & Yes\\
\hline
\cite{Bressane2015} & 5 & 540 & No \\
\hline
\cite{7880233} & 23 & 920 & No\\
\hline
AFF\cite{Fiel} & 11 & 1183 & No \\
\hline
\end{tabular}
\vspace{0.03in}
\caption{Existing bark image datasets}
\label{table:existing_datasets}
\end{table}

\subsection{Image collection and annotation}
\label{DatabaseDescription}
To solve the dataset issue, we collected images from 23 different species of trees found in parks and forests near Quebec City, Canada. We hired a forestry specialist to identify the species on site. Indeed, tree identification is much easier and reliable when relying on extra cues such as leaf shape or needle distribution. To accelerate the data collection process, we used the following protocol. First, a tree was selected and its species and circumference written on a white board by the forestry specialist. While the specialist moved to another tree, a second person took a picture of the white board as the first picture of the tree. It was then followed by 10-40 images of the bark at different locations and heights around this tree, depending on its circumference. Images were captured at a distance between 20-60~$cm$ away from the trunk. This distance was highly variable, depending on the conditions in which the photos were taken (due to obstacles, tree size, etc.). Having this kind of variability prevents overfitting to a particular distance of camera. Finally, all images were taken so as to have the trunk parallel to the vertical axis of the image plane of the camera.

We also gathered the images under varied conditions, to ensure that the dataset would be as diversified as possible. First, we used four different cameras, some of which were cellphones: Nexus 5, Samsung Galaxy S5, Samsung Galaxy S7, and a Panasonic Lumix DMC-TS5 camera. To increase the illumination variability, we took the pictures under a number of weather conditions which ranged from sunny to light rain. Finally, we selected trees from a number of different locations, such as in open areas like the university campus or parks and in the forest. This can greatly affect the appearance of the bark, especially in high vegetation density locations where the reflection of the canopy can add different shades of green to the bark color. In total, we gathered pictures during 15 outings, which took place during the summer.

\begin{figure}[ht]
\centering
\includegraphics[width=0.47\textwidth]{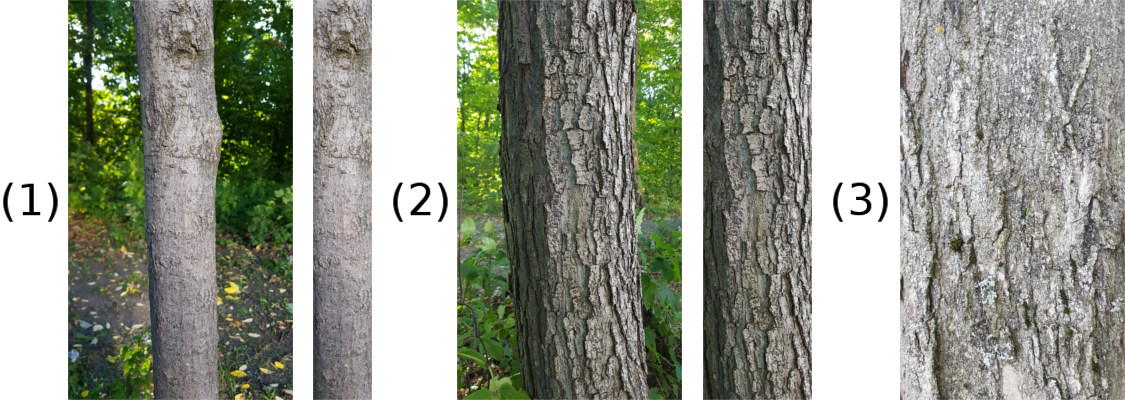}
\caption{Three images of \emph{Acer Sacharricum}, to illustrate the impact of DBH on crop size. (1) has a DBH of 8.9~$cm$, (2) has a DBH of 25.8~$cm$ and (3) has a DBH of 68.1~$cm$. Note that no cropping was needed for (3)}
\label{fig:dbh}
\end{figure}

From the picture of the white board, we obtained the species and circumference information to annotate the subsequent pictures. This means that each photo in our database contains a unique number identifying the tree, its species, its DBH, the camera used and the date and time at which it was taken. We also curated the dataset by removing approximately 25~\% of the pictures, most of them corresponding to blurred images due to camera motion. Each remaining picture was then manually cropped, so as to only keep the part of the image where bark was visible. This had the side effect that younger trees yielded very narrow pictures (Fig.~\ref{fig:dbh}~(1)), while mature trees were full-sized pictures (Fig.~\ref{fig:dbh}~(3)). Table~\ref{table:our_dataset} shows the composition of our dataset. We aimed at keeping the dataset as balanced as possible, while maximizing the number of different trees used for each class. The data collection strategy was also modulated based on initial classification results. Indeed, we increased the number of trees collected for species that were found to be difficult to separate. One can see this as a loose form of active learning, but implemented with humans in the loop. 

\begin{figure}[ht]
\centering
\includegraphics[width=0.5\textwidth]{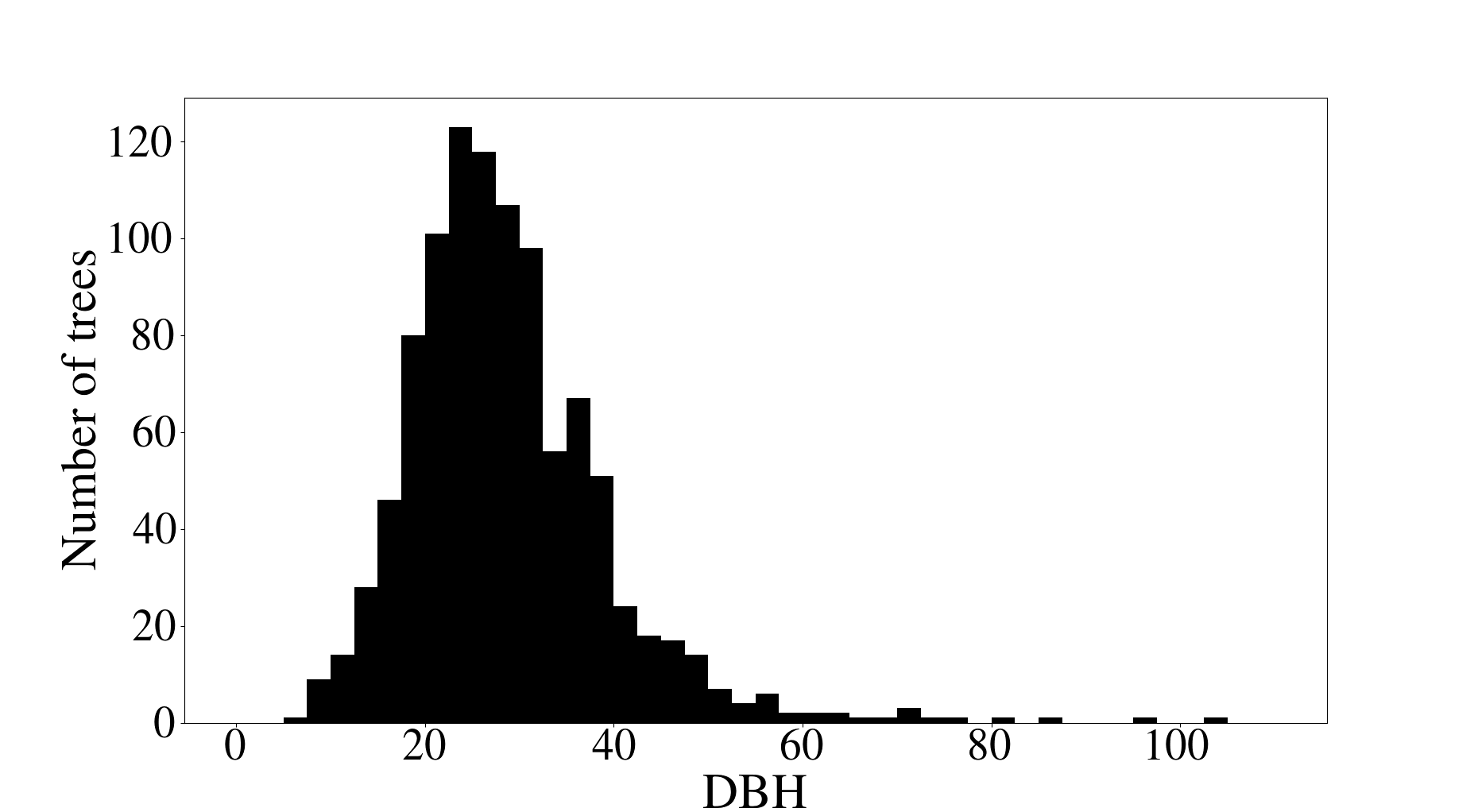}
\caption{DBH distribution of the dataset}
\label{DBH_distribution}
\end{figure}

We also aimed at having a wide distribution on the DBH which is shown in Fig.~\ref{DBH_distribution}. Most of the trees have a DBH between 20 and 30~$cm$, but we also have a few trees near 100~$cm$. This can have an impact on the classification since the size of the tree can greatly affect the appearance of the bark. Fig.~\ref{fig:dbh} shows an example of this, with the younger tree having a relatively smooth bark while the older ones are covered with ridges and furrows.

\begin{figure*}[t]
\vspace*{3mm}
\centering
\includegraphics[width=0.98\textwidth]{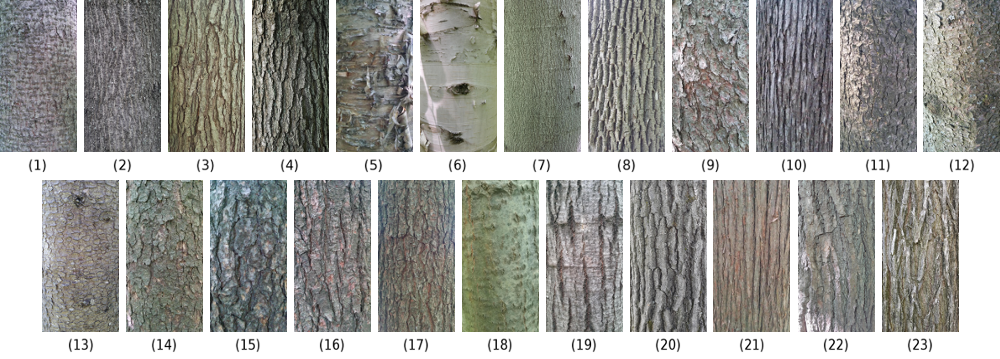}
\caption{Example image for each species. The number refers to the ID of each species, detailed in Table~\ref{table:our_dataset}. Some of the pictures have a greener tint, due to the improper white balance of the camera caused by the canopy in the forest.}
\label{fig:all_species}
\end{figure*}

\begin{table*}[t]
\begin{center}
\begin{tabular}{| c | c | c | c | c | c |}
\hline
ID & Species & Common name & Number of trees & Number of images & Number of potential unique crops \\
\hline
1 & \emph{Abies balsamea} & Balsam fir & 41 & 922 & 28235\\
2 & \emph{Acer platanoides} & Norway maple & 1 & 70 & 2394\\
3 & \emph{Acer rubrum} & Red maple & 64 & 1676 & 48925\\
4 & \emph{Acer saccharum} & Sugar maple & 81 & 1999 & 68040\\
5 & \emph{Betula alleghaniensis} & Yellow birch & 43 & 1255 & 37325\\
6 & \emph{Betula papyrifera} & White birch  & 32 & 1285 & 33892\\
7 & \emph{Fagus grandifolia} & American beech & 41 & 840 & 23904\\
8 & \emph{Fraxinus americana} & White ash & 61 & 1472 & 53995\\
9 & \emph{Larix laricina} & Tamarack & 77 & 1902 & 114956\\
10 & \emph{Ostrya virginiana} & American hophornbeam & 29 & 612 & 28723\\
11 & \emph{Picea abies} & Norway spruce & 72 & 1324 & 35434\\
12 & \emph{Picea glauca} & White spruce & 44 & 596 & 19673\\
13 & \emph{Picea mariana} & Black spruce & 44 & 885 & 43127\\
14 & \emph{Picea rubens} & Red spruce & 27 & 740 & 22819\\
15 & \emph{Pinus rigida} & Pitch pine & 4 & 123 & 2264\\
16 & \emph{Pinus resinosa} & Red pine & 29 & 596 & 14694\\
17 & \emph{Pinus strobus} & Eastern white pine & 39 & 1023 & 25621\\
18 & \emph{Populus grandidentata} & Big-tooth aspen & 3 & 64 & 3146\\
19 & \emph{Populus tremuloides} & Quaking aspen & 58 & 1037 & 63247\\
20 & \emph{Quercus rubra} & Northern red oak & 109 & 2724 & 72618\\
21 & \emph{Thuja occidentalis} & Northern white cedar & 38 & 746 & 19523\\
22 & \emph{Tsuga canadensis} & Eastern Hemlock & 45 & 986 & 27271\\
23 & \emph{Ulmus americana} & American elm & 24 & 739 & 27821\\
\hline
Total & - & - & 1006 & 23616 & 817647\\
\hline
\end{tabular}
\end{center}
\caption{\emph{BarkNet 1.0} dataset composition. Although we used random crops during training, we indicate in the last column the number of unique crops (without any overlap) that can be theoretically generated for training purposes}
\label{table:our_dataset}
\end{table*}

\section{Experiments} \label{sec:experiments}

\subsection{Architecture}
As is commonly done in image recognition tasks, we employed networks that have been pre-trained on ImageNet. Moreover, we used the ResNet architecture \cite{He2015}, as it is both powerful and easy to train on standard classification problems.

\subsection{Training Details}
We used PyTorch \texttt{0.3.0.post4}~\cite{pytorch} for all experiments and downloaded the weights of the \texttt{resnet18} and \texttt{resnet34} networks pre-trained on ImageNet. As commonly-accepted practice, we froze the first layer, since our problem is very different from ImageNet, and then fine-tuned the networks using an initial learning rate of 0.0001. We reduced the learning rate at fixed epochs (16 and 33) by a factor of 5, and trained for a total of 40 epochs. We used Adam as the optimization method, with a weight decay of 0.0001.

Since the photos are high definition, we resized them to half of their original size. This allowed for a faster loading and image processing of the images when creating the mini-batches. It also takes into account the Bayer filter pattern on color cameras, which only samples colors for every other pixel on the imaging element. For each mini-batch, we uniformly sampled a random tree species (class), from which we sampled a random image from a random tree. This allowed us to mitigate the problems of having an unbalanced dataset, similarly to the class-aware sampling used in~\cite{RelayBackpropECCV2016}. Then, we augmented the data using random horizontal flips and finally, we took a random crop of 224x224 pixels in the resulting image. Recall that during the data gathering process, a fair amount of randomness in terms of illumination and scale was present, so we did not perform color, scale or contrast jittering.

\section{Results} \label{sec:results}

In our experiments, we compared the effect of network depth (18 vs 34) on classification precision. We also tested for different batch sizes, to evaluate its regularization effect~\cite{FlatMiniBatch2018}. For the evaluation, we used a 5-fold cross-validation method using 80\% of the trees for the training and the remaining for testing. Care was taken in performing the split on the trees instead of the image, to avoid positively biasing results due to the network learning to recognize individual trees instead of the species.
We report the average accuracy on the 5 folds for two different scenarios: (i) one where we evaluate all the image individually as if they were all from different trees and (ii) one where we classify each tree by using all of its images.
Note that we did not use \emph{Acer platanoides}, \emph{Pinus rigida} and \emph{Populus grandidentata} since we did not collect enough images in these categories to obtain meaningful results.

\subsection{Test results when using individual images}
\label{SinglePictureTest}
\begin{figure}[t]
\vspace*{3mm}
\centering
\includegraphics[width=0.47\textwidth]{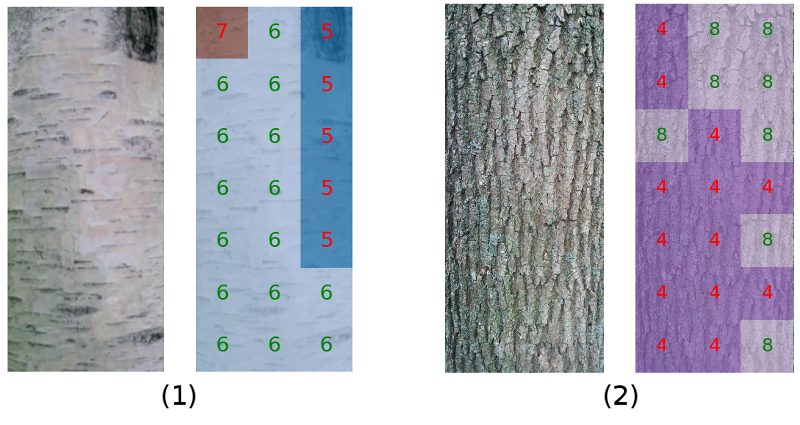}
\caption{Examples of multiple crop majority classification. (1) shows a successful classification on \emph{Betula papyrifera} and (2) shows a classification error on \emph{Fraxinus americana}. The green number indicates the correct class and red numbers indicate the incorrect classes. The numbers refer to the ID of each species, detailed in Table \ref{table:our_dataset}}
\label{fig:crop_exemple}
\end{figure}

Table~\ref{table:results_images} contains the results of evaluating the two models on each image individually, for a number of batch sizes. We report both single crop (random) and multiple crop results. For the latter, we split the test image into multiple non-overlapping 224x224 crops and classified each one individually. Then, we performed majority voting to determine the final outcome. As can be seen from Table~\ref{table:results_images}, progressing from single crops (87.04\%) to multiple crops (93.88\%) on a complete image significantly improves the accuracy, which is expected. Fig.~\ref{fig:crop_exemple} displays two examples of classification using the multiple tiled crops, showing the spatial distribution of the classification. It also displays the ID label for each crop.

\begin{figure*}[t]
\vspace*{3mm}
\centering
\includegraphics[width=0.98\textwidth]{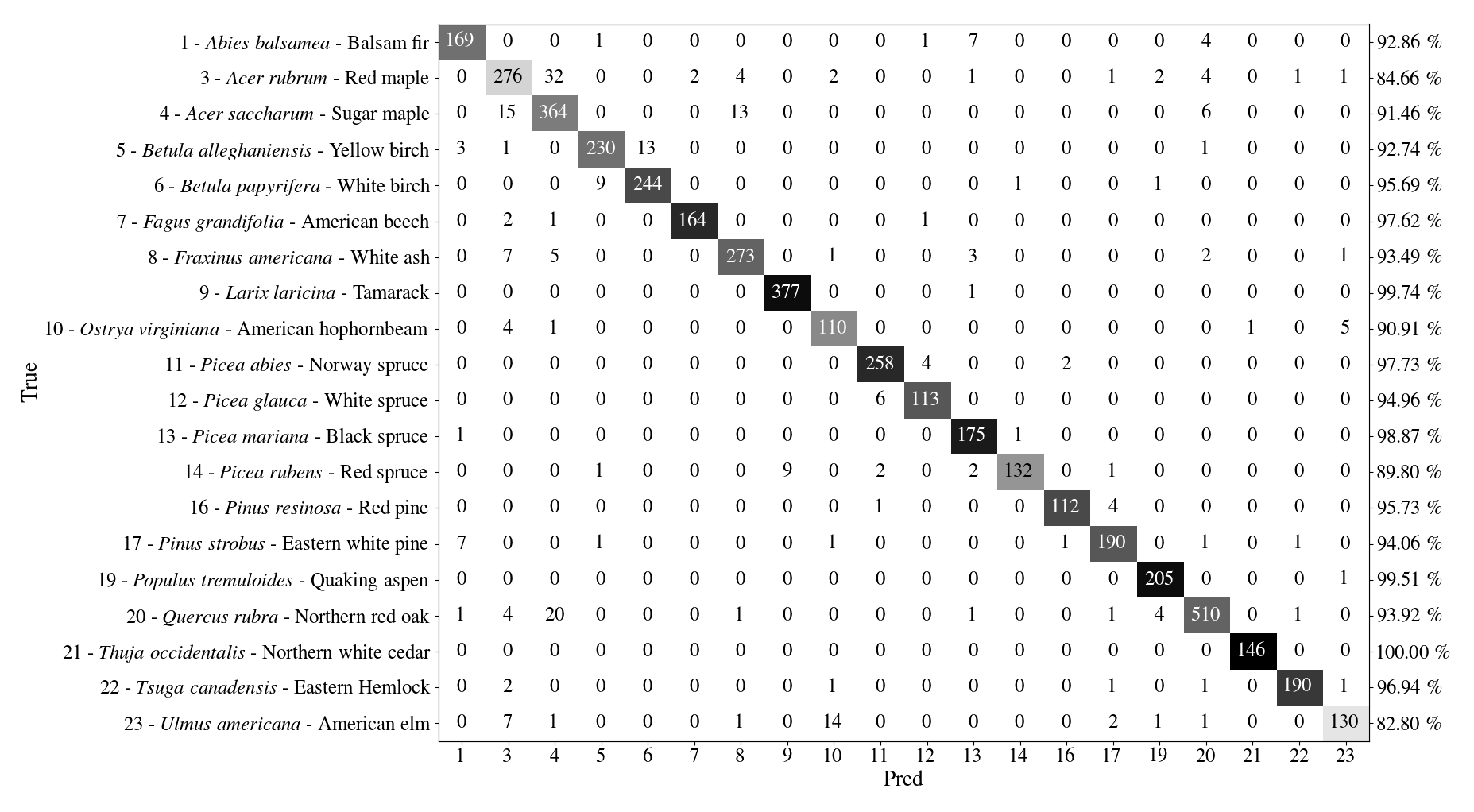}
\caption{Average confusion matrix for multiple crops voting on whole images using a \texttt{resnet34} and a batch size of 32}
\label{fig:confusion_matrix}
\end{figure*}

Fig.~\ref{fig:confusion_matrix} shows the average confusion matrix of our multiple crops voting on individual image experiments using a \texttt{resnet34} and a batch size of 32. As one may suspect, trees from the same family are more difficult to differentiate. For instance, \emph{Betula parpyrifera} and \emph{Betula alleghaniensis} as well as \emph{Acer rubrum} and \emph{Acer saccharum} are often confused with one another. Fig.~\ref{fig:confusion_matrix} also shows some other difficult combinations, such as \emph{Fraxinus americana} and \emph{Acer saccharum}.

\begin{table}[t]
\begin{center}
\begin{tabular}{| c | c | c | c |}
\hline
\multicolumn{4}{| c |}{Images}\\
\hline
Network & Batch size & Single crop & Multiple crops\\
\hline
\multirow{ 4}{*}{\texttt{resnet18}} & 8 & 85.00 & 92.93\\
& 16 & 85.89 & 93.16\\
& 32 & 85.87 & \textbf{93.23}\\
& 64 & \textbf{86.03} & 93.04\\
\hline
\multirow{ 4}{*}{\texttt{resnet34}} & 8 & 85.24 & 93.15\\
& 16 & 86.75 & 93.50\\
& 32 & \textbf{87.04} & \textbf{93.88}\\
& 64 & 86.96 & 93.43\\
\hline
\end{tabular}
\end{center}
\caption{Classification accuracy on the \emph{individual image} scenario, for different batch sizes. Single crop and multiple crops results are reported}
\label{table:results_images}
\end{table}

\begin{table}[t]
\vspace*{3mm}
\begin{center}
\begin{tabular}{| c | c | c | c |}
\hline
\multicolumn{4}{| c |}{Trees}\\
\hline
Network & Batch size & Single crop & Multiple crops\\
\hline
\multirow{ 4}{*}{\texttt{resnet18}} & 8 & 97.10 & 97.51\\
& 16 & 97.21 & 97.50\\
& 32 & 96.92 & 97.21\\
& 64 & \textbf{97.31} & \textbf{97.70}\\
\hline
\multirow{ 4}{*}{\texttt{resnet34}} & 8 & \textbf{97.81} & 97.22\\
& 16 & 97.50 & 97.11\\
& 32 & 97.50 & \textbf{97.51}\\
& 64 & 97.30 & 97.31\\
\hline
\end{tabular}
\end{center}
\caption{Classification accuracy on \emph{all images per tree} scenario, for different batch sizes. Single crop and multiple crops results are reported}
\label{table:results_tree}
\end{table}

\subsection{Test results when using all images of a tree}
We were interested in seeing if the use of images taken at several different locations along the trunk would improve the classification results. We thus performed majority voting across all of the images of a given tree, both for single and multiple crops per pictures. Note that the number of available images per tree was variable, as stated in Section~\ref{DatabaseDescription}. Table~\ref{table:results_tree} contains the results of this evaluation, again for a number of batch sizes. The results indicate that we are able to further improve the classification results (97.81\%). More interestingly, we did not see any real difference between using a single or multiple crops in each image. This seems to indicate that having a greater variety of locations along a trunk is more beneficial than having a large number of crops that are closely located. This can probably be explained by anecdotal observations in the field, where we noticed that the bark appearance changed significantly from one trunk region to another.

\subsection{Effect of dataset size on training performance}

\begin{figure}[t] 
\centering
\includegraphics[width=0.47\textwidth]{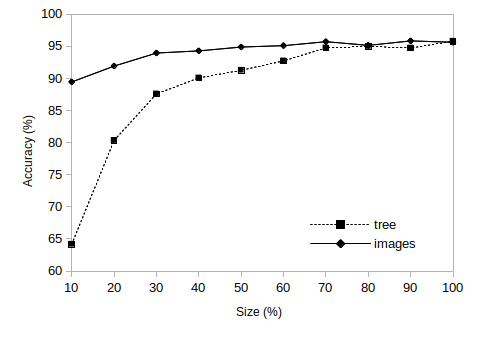}
\caption{Results obtained with a \texttt{resnet34} when only using a smaller percentage of \emph{a)} images or \emph{b)} trees from the dataset with multiple crops per image}
\label{fig:dataset_size}
\end{figure}

A common question arising when developing new classifier systems is: how much data do we need for training purposes? To answer this, we empirically evaluated the impact of the size of the training dataset on the classification accuracy. Moreover, we performed this evaluation for two cases that are particular to our classification problem: \emph{a)} reduced number of images and \emph{b)} reduced number of individual trees. To accomplish this, one fold from the previous experiment in Section~\ref{SinglePictureTest} was taken and 9 smaller training datasets were created from the training set per case. For case \emph{a)}, we randomly sampled images from the training set until we hit a target goal of images. For case \emph{b)}, instead of sampling the images, we sampled the individual trees directly until we hit a target number of trees. Fig.~\ref{fig:dataset_size} shows the results obtained. Note that we used the same testing set in both cases.

As can be seen, the general trend is that an increase in the number of images for the training leads to better results. However, the network is much more sensitive to the number of trees in the training dataset, rather than to the overall number of pictures. Indeed, when the number of overall images is randomly reduced by 90\%, only about 5\%
of accuracy is lost. On the other hand, when the number of trees is randomly reduced by 90\%, the results fall by more than 30\%. This indicates that it is much more important to collect training data over a large number of trees, rather than taking a large number of pictures per tree. In other words, only a fairly limited number of pictures per tree are required to obtain a good performance.

\section{Conclusion and future work} \label{sec:conclusion}

In this paper, we have empirically demonstrated the ability for ResNets to perform tree species identification from the pictures of bark, for 20 Canadian species. On our collected dataset, the accuracy of the method ranges from 93.88\% (for multiple crops on a single image) to 97.81\% (using all trunk images), far above the 5\% chance classification. We have found empirically that training is significantly more susceptible to the number of trees in our database rather than the overall number of images. This result will help tailor further data gathering efforts on our side.

In the process, we have also created a large public dataset (named \emph{BarkNet} 1.0) containing labeled images of tree barks. This database can be used to accelerate research on bark classification for robotics or forestry applications. It can also contribute in helping the computer vision community develop algorithms on the challenging problems of fine-grained texture classification. 

Nevertheless, more work is needed to adapt the architecture of the network specifically to this task. As future work, we aim to leverage the DBH into a multi-task approach~\cite{Trottier2017}. The use of multi-scale classifications will also be studied in an effort to determine the optimal scale at which to perform bark image classification. Moreover, we will explore the use of novel deep architectures that have been tailored to texture classification. We also plan on testing the approach on a sawmill floor, where we will have access to thousands of logs for data gathering. A new challenge will be to ensure that damages to bark due to logging operations do not adversely affect classification performances.

\addtolength{\textheight}{-8.4cm}   




\section*{ACKNOWLEDGMENT}

The authors would like to thank Luca Gabriel Serban and Martin Robert for their help in creating this dataset.


\bibliographystyle{IEEEtran}
\bibliography{references}

\end{document}